\documentclass[mathematics,article,accept,pdftex,moreauthors]{Definitions/mdpi} 
\usepackage[ruled,linesnumbered]{algorithm2e}
\usepackage{changes}
\firstpage{1} 
\pubvolume{1}
\issuenum{1}
\articlenumber{0}
\pubyear{2022}
\copyrightyear{2022}
\externaleditor{Academic Editor:{Weihua Xu} 
}
\datereceived{20 November 2022} 
\dateaccepted{13 December 2022} 
\datepublished{} 
\hreflink{https://doi.org/} 
\Title{Automatic Semantic Modeling for Structural Data Source with the Prior Knowledge from Knowledge Base $^{\dagger}$}

\TitleCitation{Automatic Semantic Modeling for Structural Data Source with the Prior Knowledge from Knowledge Base}

\Author{Jiakang Xu 
 $^{1,2,3,4}$, Wolfgang Mayer $^{5}$, HongYu Zhang 
 $^{2,3,4,}$*, Keqing He $^{6}$ and Zaiwen Feng $^{1,2,3,4,7,8,}$*}

\AuthorNames{Jiakang Xu, Wolfgang Mayer, Hong-Yu Zhang, Keqing He, Zaiwen Feng}

\AuthorCitation{Xu, J.; Mayer, W.; Zhang, Ho.; He, K.; Zhang, J.; Feng, Z.}

\address{%
$^{1}$ \quad National Key Laboratory of Crop Genetic Improvement, Huazhong Agricultural University, \linebreak  Wuhan 430070, China; xjk@webmail.hzau.edu.cn \\
$^{2}$ \quad Hubei HongShan Laboratory, Huazhong Agricultural University, Wuhan 430070, China\\ 
$^{3}$ \quad College of Informatics, Huazhong Agricultural University, Wuhan 430070, China\\
$^{4}$ \quad Hubei Key Laboratory of Agricultural Bioinformatics, Huazhong Agricultural University, \linebreak  Wuhan 430070, China \\
$^{5}$ \quad Industrial AI Research Centre, University of South Australia, Mawson Lakes, SA 5095, Australia; wolfgang.mayer@unisa.edu.au \\
$^{6}$ \quad School of Computer, Wuhan University, Wuhan University, Wuhan 430072, China; hekeqing@whu.edu.cn \\
$^{7}$ \quad State Key Laboratory of Hybrid Rice, Wuhan University, Wuhan 430072, China \\
$^{8}$ \quad Macro Agricultural Research Institute, Huazhong Agricultural University, Wuhan 430070, China \\
}

\corres{Correspondence:   zaiwen.feng@mail.hzau.edu.cn (Z.F.) and zhy630@mail.hzau.edu.cn (H.Y.Z.)}

\firstnote{This paper is an extended version of our paper published in 2021 IEEE International Conference on Data, Information, Knowledge and Wisdom (DIKW), Haikou, Hainan, China, 20--22 December 2021; pp. 2034--2041.} 

\abstract{A critical step in sharing semantic content online is to map the structural data source to a public domain ontology.  This problem is denoted as the Relational-To-Ontology Mapping Problem (\textit{Rel2Onto}). A huge effort and expertise are required for manually modeling the semantics of data. Therefore, an automatic approach for learning the semantics of a data source is desirable. Most of the existing work studies the semantic annotation of source attributes. However, although critical, the research for automatically inferring the relationships between attributes is very limited. In this paper, we propose a novel method for semantically annotating structured data sources using machine learning, graph matching and modified frequent subgraph mining to amend the candidate model. In our work, \textit{Knowledge graph} is used as prior knowledge. Our evaluation shows that our approach outperforms two state-of-the-art solutions in tricky cases where only a few semantic models are known.}


\begin{document}




\section{Introduction}

Structural data sources are still one of the most prevalent modes for the storage of enterprise or Web data. It is a long-standing and urgent issue in many real database research fields to automatically integrate heterogeneous data {sources} 
 \cite{ref_article1,ref_article2}. Though relational schemata are suitable for ensuring data integrity, they are short of semantic descriptions that support efficient semantic integration of different sources. {The construction of ontology from relational databases is a basic problem for the development of the Semantic Web \cite{ref_article34}.} Common ontology provides a method to express the semantics of a relational schema and facilitate integrating heterogeneous data sources. It is appropriate to manually indicate semantic descriptions if the integration of only a few data sources is required. However, as the number of heterogeneous schemata increases, manually labeling becomes tedious. To settle this problem, a standard way is to design a common ontology and build a source description of the assigned mappings between the sources and the ontology \mbox{automatically \cite{ref_article3}.} This problem is named as Relational-To-Ontology Mapping Problem (\verb|Rel2Ont|) \cite{ref_article11}.

Formally, the problem of \verb|Rel2Ont| can be expressed as below \cite{ref_article11}. Assume a data source $s$ is an \textit{n}-ary relation which contains a series of attributes $\mathcal{A}_s=(a_1,...,a_n)$.  The \textit{attribute mapping} function $\phi : \mathcal{A}_s \mapsto \mathcal{D}_m$ constructs the mapping between the attributes of the sources $s$ and the node set $\mathcal{D}_m$ of the semantic model $m$. Given a source $s$, a semantic model $m$, and an attribute mapping $\phi$, a source description is defined as a triple $\delta = (s,m,\phi)$. {Suppose we have a gold standard model for a new source $s^*$}, given an ontology $\mathcal{O}$, and a set of source descriptions $\Delta=\{(s_1,m_1,\phi_1),...,(s_l,m_l,\phi_l)\}$ , for a new source $s^*$.  How do we derive the semantic model $m^*$ and the attribute mapping function $\phi^*$ so that $\delta^*=(s^*,m^*,\phi^*)$ maximizes the \textit{precision} and \textit{recall} between the semantic model $m^*$ and the \textit{gold standard} semantic model $m^\dagger$ of the data source $s^*$?

Lately, machine learning techniques have been employed to address the \verb|Rel2Ont| issue. For instance, Karma \cite{ref_article4} can automatically learn semantic models for a new data source by leveraging the knowledge from domain ontology and historical semantic models of sources in the same domain. Binh Vu {et al.} \cite{ref_article27} proposed a novel way to learn semantic models for data sources by using a probabilistic graphical model (PGM). Recently \textit{knowledge graphs}, as one of the main trends driving the next wave of technologies \cite{ref_article21}, have become a novel form for representing knowledge and the basis of multiple applications from common applications to specific industrial use cases \cite{ref_article22}. {A knowledge graph is a structured representation of facts consisting of entities, relationships, and semantic descriptions \cite{ref_article31}.} Relationships among semantic models of a data source can be inferred by exploiting a knowledge graph as prior knowledge \cite{ref_article5}. {For example, Giuseppa {et al.} developed a semi-automatic tool SeMi for constructing large-scale knowledge graphs from structured data sources by building the semantic models of data sources \cite{ref_article35}. }

These automatic semantic annotation methods based on machine learning enormously elevate the data matching efficiency of structural data sources. However, these strategies also have the following drawbacks: (a) Limited known semantic models of data sources: the performance of \verb|Karma| is better when plenty of data sources are available for training a standard learning graph. For instance, when \verb|Karma| uses 29 known museum data sources to train the learning graph, it can generate a candidate semantic model for the new museum data source with an accuracy of up to 80$\%$ \cite{ref_article4}, whereas in practical applications, the data sources might be limited, perhaps to 2 or 3, so the existing methods still need to be greatly improved. (b) Lack of linked data: when inferring semantic relations from knowledge graphs based on machine learning methods \cite{ref_article5}, it is assumed that there are adequate linked data usable in the same domain as the destination data source, which greatly depends on the amount of linked data. If there is little or no linked data available, the capacity to dig available patterns will be greatly reduced. (c) Finite ability to extract long patterns: the calculation of inferring patterns from the method of inferring semantic relationships from knowledge graphs is very complicated \cite{ref_article5}. In a reasonable time, only the patterns of three or four  nodes may be inferred. Therefore, it is  challenging to use SPARQL queries to extract long patterns from a lot of triples.

In this article, we extend our previous work \cite{ref_article30} and present a novel machine-learning-based procedure, which is helpful to figure out the \verb|Rel2Onto| issue with the prior knowledge of the knowledge graph. For this reason, we attribute the \verb|Rel2Onto| problem to a customized frequent subgraph mining problem. Primarily, it runs the Steiner tree generation algorithm to output reasonable semantic models by utilizing existing semantic models and domain ontology \cite{ref_article4}. We select the first ranked candidate model \textit{cm} as the seed model for further amending. Then, we remove incorrect relationships of \textit{cm} by using a knowledge graph as prior knowledge and machine learning techniques. Since some relationships are removed from the initial candidate semantic model, the resulting model may be incomplete.  Accordingly, to improve the completeness of the model, we use a \textit{grow-and-store} strategy \cite{ref_article6} to mine the top-$\sigma$ frequent subgraphs in the knowledge graph as final candidate models. The underlying \emph{heuristic} hypothesis is that the correct semantic model may have higher frequency in the knowledge graph than other substructures.

The contributions of our paper are as follows: (a) A new pipeline for automatically learning the semantic model of a new structural data source is arranged by utilizing several existing semantic models, domain ontologies, and domain specific knowledge graphs. (b) A novel approach is put forward for discovering and eliminating the incorrect relationships in candidate semantic models by machine learning and graph matching techniques. (c) We use the the \textit{grow-and-store} approach and modify a frequent subgraph mining algorithm to calculate the subgraph frequency of a domain-specific knowledge graph and mine frequent subgraphs. The top-$\sigma$ frequent subgraphs are acquired as the most rational semantic models of the structured data source.

The rest of this article is structured as below. In Section \ref{22}, we represent related work. We introduce an illustrative example in Section \ref{33}. We exploit a new way for inferring semantic relations of the destination data source in Section \ref{44}. In Section \ref{55}, we display our experimental evaluation results of our method and draw conclusions in Section \ref{66}.

\section{Related Work}\label{22}
Since the manual creation and scheme of the mapping between relations to ontologies is a labor-intensive process, several \emph{machine learning} technologies have been suggested to tackle the \verb|Rel2Ont| problem. Taheriyan {et al.} presented a method to automatically learn a semantic model of a new source utilizing domain information and historical semantic models \cite{ref_article4}. Considering lack of known available semantic model in many domains, Taheriyan {et al.} further put forward a method to automatically learning the semantic relationships within a given data source exploiting Linked Open Data (LOD) \cite{ref_article5}. The limitation of \cite{ref_article5} is that when the available LODs are sparse, the exactness of the outputted semantic models is severely affected and not enough useful patterns are obtained. Diego {et al.} \cite{ref_article11} combined machine learning with constraint programming to infer mapping rules from previous mapping instances to deal with attributes that cannot be matched to the ontology. Binh {et al.} \cite{ref_article27} presented a method for automatically learning semantic models using a probabilistic graphical model. Their approach is more robust to noise than previous methods. {Giuseppe {et al.} proposed a semi-automatic approach for inferring semantic relations based on a graph neural network trained on a background knowledge graph \cite{ref_article35}.} In our article, we learned a semantic model for a new source by integrating the knowledge of known semantic mappings, knowledge graphs, and domain ontology. Our approach helps to study the exact semantic models without sufficient historical mappings. We studied this problem in our previous work \cite{ref_article30}. Comparing to our previous work, the following four points are strengthened in this article. First, multiple candidate semantic types for each attribute of structural data source are considered, while in our previous work, we assumed that all the correct semantic types of attributes are known. Second, we eliminated the incorrect relationships in \textit{seed model} by using two different methods in the pipeline to improve the accuracy of our method. Third, we optimized our previous algorithm for adding missing substructures to improve its efficiency. Finally, we conducted more experiments to evaluate our approach by using more datasets. Compared to our previous work with only one dataset and one semantic modeling approach being and evaluated, in this article, we compared our approach with two state-of-the-art semantic modeling methods by exploiting three datasets.

\emph{Search}-based approaches were also exploited to tackle the \verb|Rel2Ont| problem. Pinkel {et al}. proposed a new semi-automatic matching method \textit{IncMap} \cite{ref_article16}, which used the lexical and structural similarities between ontology and relational schemata to generate the mapping from relational schemata to ontology. Sequeda {et al}. defined a specific query mapping QODI \cite{ref_article17} for an ontology-based data integration system (OBDI). QODI generates path correspondences , rather than entity correspondences, to facilitate the representation of queries. Moreover, Sequeda {et al.} exploited a semi-automatic software Ultrawrap Mapper \cite{ref_article18} for the creation of a mapping from Relational Databases to RDF in the R2RML language. Ultrawrap Mapper aligns the original schemata and target ontology using QODI techniques and gives mapping suggestions \cite{ref_article18}. The MIRROR system \cite{ref_article19} yields an R2RML mapping file containing the mappings for a given relational database. {Naglaa {et al}. proposed a ProGOMap (Property Graph to Ontology Mapper) system for the automatic generation of mappings from property graphs to a domain ontology \cite{ref_article32}. The PG-to-Ontology mappings can be automatically generated by using the aligned axioms.} {Florian {et al}. developed a prototype for a semantic data lake for addressing the heterogeneous format of the activity logs and the content data of cross-platform collaboration \cite{ref_article33}. In their prototype, ontology-based data access is implemented based on a mapping between an ontology and the ingested data}. All of the approaches in \cite{ref_article16, ref_article17, ref_article18, ref_article19, ref_article32, ref_article33} focus on constructing mappings between relational scheme and domain ontology through a search-based algorithm. Different from these search-based approaches, our method attempts to learn mapping rules by leveraging the amount of knowledge bases, including not only historical relational schemes and domain ontology, but also knowledge graphs.

As mentioned above, \emph{semantic labeling} is a significant step for solving the \verb|Rel2Ont| problem.  Several works for addressing the problem of semantic labeling exist. For instance, Krishnamurthy {et al.} \cite{ref_article23} tried to leverage the distribution and characteristic properties of the data for learning semantic types of source attribute. Pham {et al.} \cite{ref_article7} used machine learning technologies to infer the correct semantic type by calculating similarities between unlabeled and labeled attributes.  Mulwad {et al.} \cite{ref_article8} leveraged Linked Open Data (LOD) information to build up the known semantic message algorithm for the annotation of tables on the web.

\section{Illustrative Example}\label{33}

\begin{figure}[H]
\begin{adjustwidth}{-\extralength}{0cm}
\centering
\includegraphics[width=1.2\textwidth]{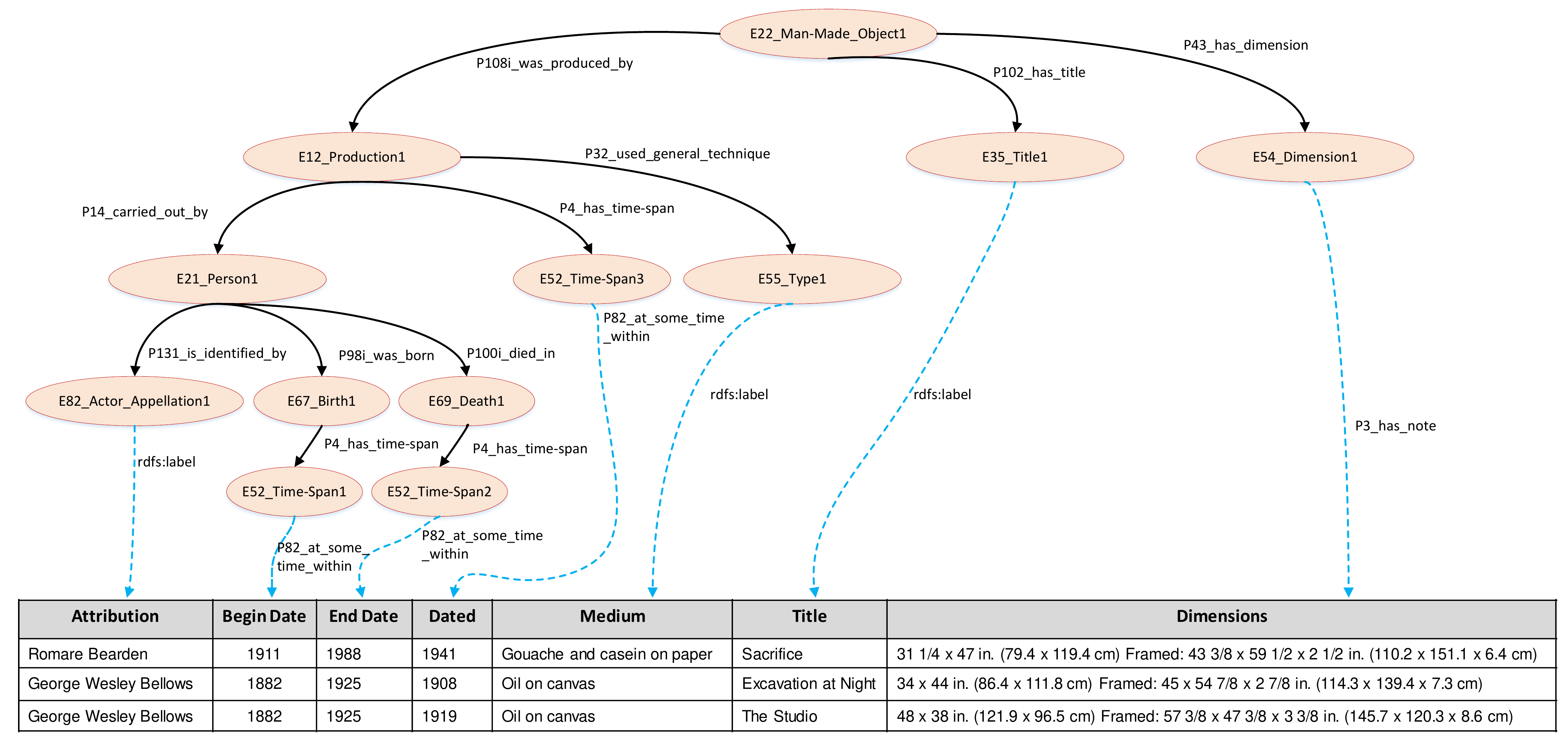}
\end{adjustwidth}
\caption{CB data source and its semantic model.} 
\label{cb_data source and its semantic model}
\end{figure}

In this section, a typical procedure for building a semantic model of a sample data source from the Crystal Bridges Museum in the {United State~(CB)} 
 {\url{https://crystalbridges.org/ }}(accessed on 14 December 2022) {is provided.} 
 The gold semantic model of the CB is exhibited in Figure~\ref{cb_data source and its semantic model}. In the article, the \textit{correct} model of a data source means the gold standard model of this source.

The semantic model $m$ is a directed graph which includes two categories of nodes: \textit{class nodes} ($\mathcal{C}_m$) and \textit{data nodes} ($\mathcal{D}_m$). $\mathcal{C}_m$ is the set of classes in the ontology, and $\mathcal{D}_m$ corresponds to data properties. {The term semantic model is synonymous with semantic mapping with a minor difference. Semantic mapping is a schema mapping from the data source to an ontology. This mapping can be represented as a semantic graph which is also called a semantic model.} As Figure~\ref{cb_data source and its semantic model} shows, the class nodes are ontology classes, and the data nodes are source attributes. For example, in the seed model of CB, the entities \verb|E55_Type1| and \verb|E54_Dimension1| are class nodes, and data source attributes \verb|Medium| and \verb|dimensions| are data nodes. The edges of the semantics model can be  classified into \textit{object properties} and \textit{data properties}. In Figure~\ref{cb_data source and its semantic model}, object properties defined in the ontology are shown by the black-colored links between class nodes. Furthermore, data properties (shown in blue color) are relationships between data nodes and class nodes. For instance, \verb|P102_has_title| and \verb|rdfs:label| are object properties and data links, respectively.

Suppose that we only have two data sources whose gold standard semantic models are given. The data sources are the tables describing the information about artworks in the National Portrait Gallery (NPG) and Getty: Resources for Visual Art and Cultural Heritage (GT), respectively. Our goal is to learn a semantic model for a new source such as CB by leveraging a small number of historical semantic mappings and the domain ontology and using a knowledge graph as background knowledge. In the next section, we present our method for automatically learning a semantic model for a new source with a few known semantic models and a knowledge graph.

\section{Our Approach}\label{44}

\begin{figure}[H]
	\includegraphics[width=\textwidth]{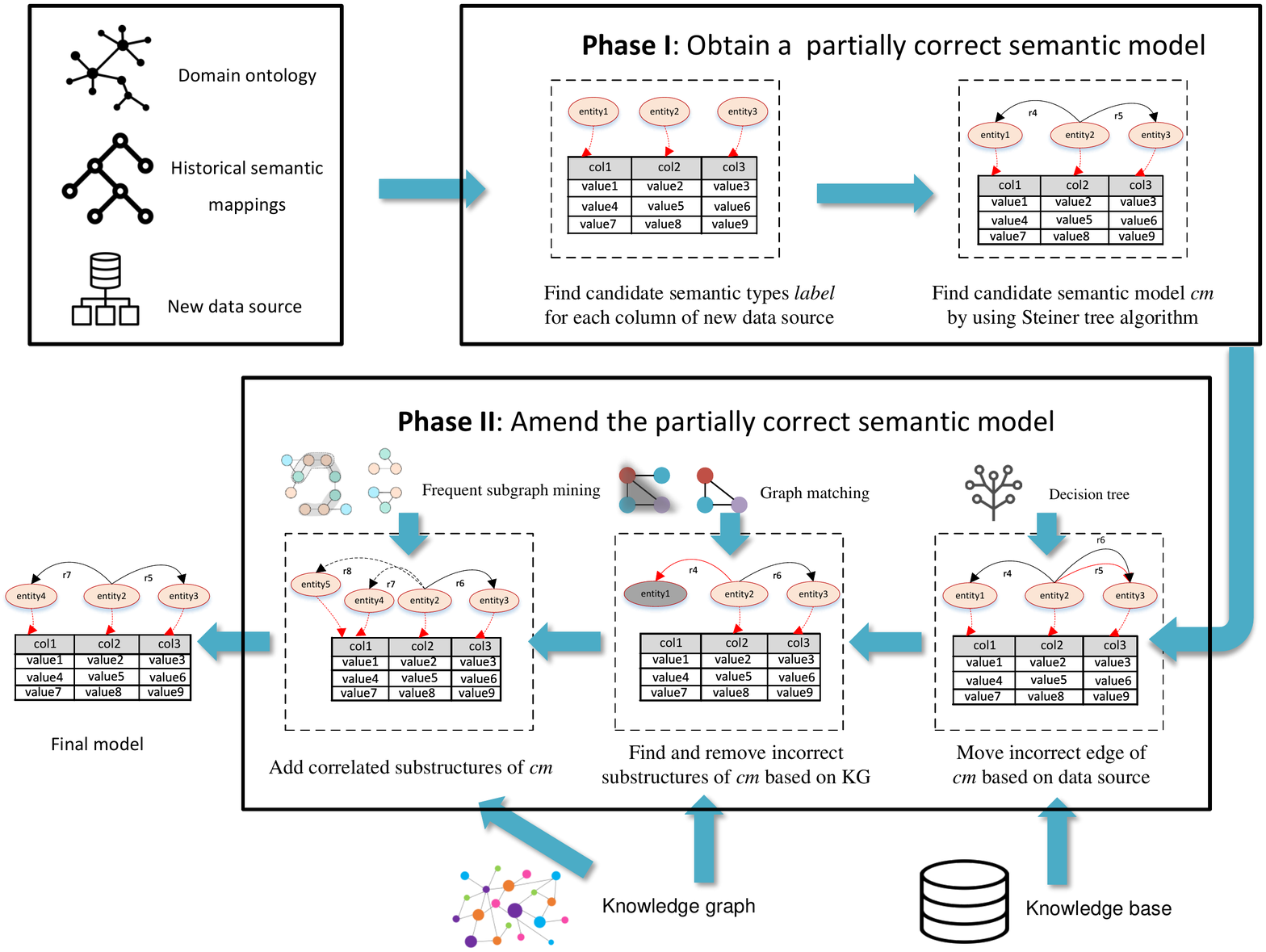}
\caption{The overall pipeline.} 
\label{overall_pipeline}
\end{figure}

In this section, we introduce our method for automatically inferring the semantic models of a structural data source. The whole pipeline of this method is illustrated in Figure~\ref{overall_pipeline}. The domain ontology, several historical semantic mappings from data sources to domain ontology, a new data source, and domain knowledge graphs are the inputs of our approach. The overall pipeline consists of two phases. In the first phase, we find candidate semantic types for each attribute and find a candidate semantic model {cm} 
 for a new data source by using a Steiner tree algorithm. A cm is usually  
a partially correct model. In the second phase, we amend the cm to eliminate incorrect entities and relationships. We first train a decision tree model to distinguish ambiguous relationships and then use a graph matching technique to remove incorrect relationships. Next, a modified frequent subgraph mining algorithm is used to add missing substructures. The output of our approach is a semantic model that describes how the specified semantic types are linked with the \mbox{seed model.} 

\subsection{Obtaining the Seed Semantic Model}

The first phase abides by the classical solution \cite{ref_article3} of the \verb|Rel2Ont| problem which includes two sub-steps, i.e., \emph{semantic labeling} and \emph{relationship discovery}. Semantic labeling \cite{ref_article10} is the process of annotating the semantic type for an attribute by scoring a confidence value for the assignment of an attribute from $s$ to a type $l \in L_o$ {($L_o$ is the set of all possible candidate semantic types)}. In this work, we use the approach proposed by Krishnamurthy {et al.} \cite{ref_article23} which is called {SemanticTyper} to automatically obtain the semantic type of source attributes. Semantic types for each attribute of a new data source can be labeled automatically by using a semantic labeling function trained from a set of manually annotated sources. 

\begin{figure}[H]
\begin{adjustwidth}{-\extralength}{0cm}
\centering
\includegraphics[width=1.2\textwidth]{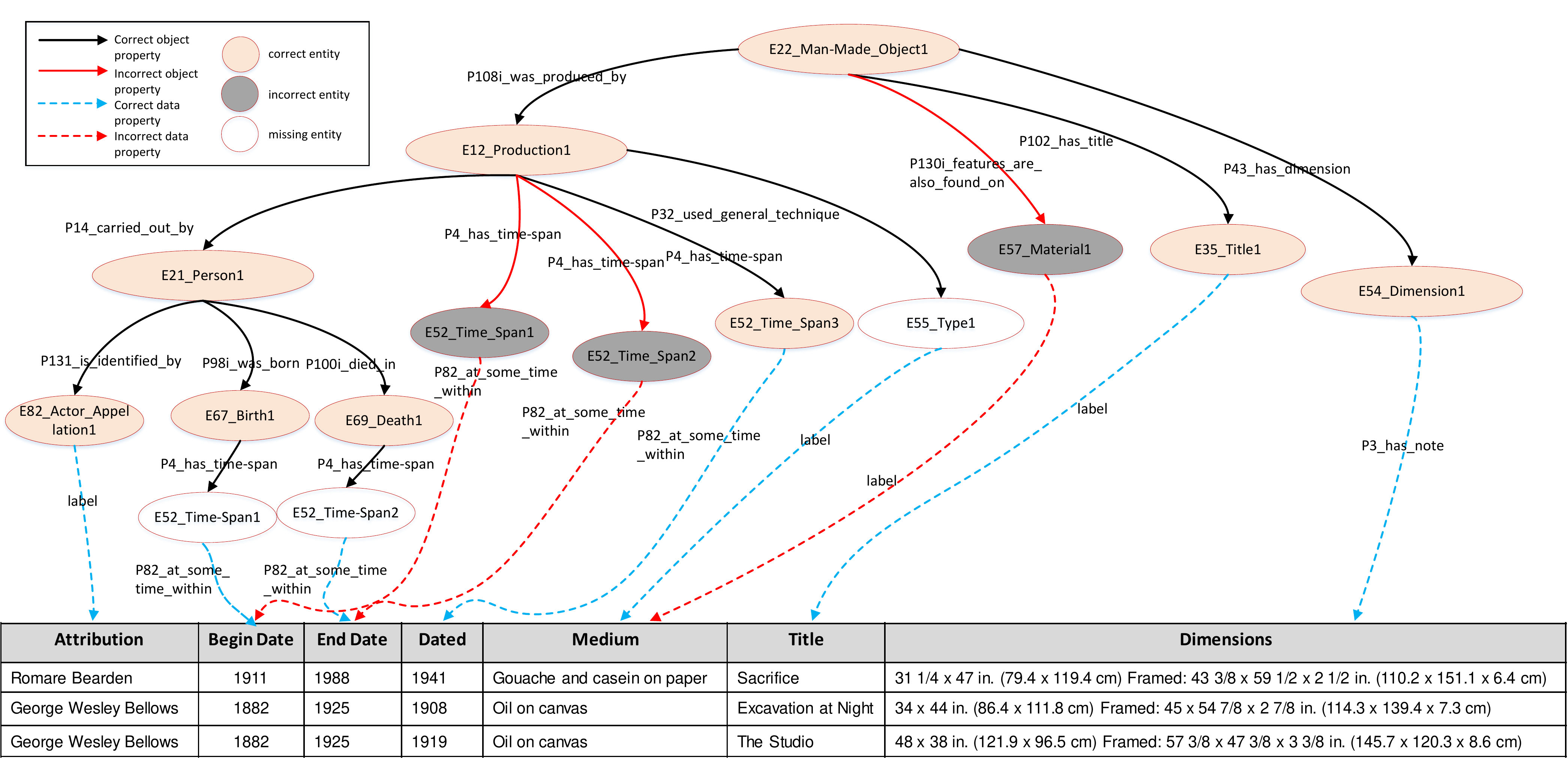}
\end{adjustwidth}
\caption{The seed model of CB. The incorrect entities are in the gray background, and incorrect relationships are colored red. The missing entities are on a white background} 
\label{seed}
\end{figure}

Relationship discovery is the process of linking all the semantic annotations with multiple relationships to formulate a semantic model. We find all the relationships and build a candidate model cm by using a Steiner tree algorithm. The relationships of the matched data nodes (attributes) are identified for the generation of the semantic models $T^*=\{T^1,T^2,...,T^k\}$ of data sources. First, a directed weighted graph $\mathcal{G}_\mathcal{O}=(\mathcal{V}_\mathcal{O},\mathcal{E}_\mathcal{O})$, called \textit{alignment graph}, is constructed on top of the historical semantic mappings and expanded using semantic types $\mathcal{L}_\mathcal{O}$ and the ontology $\mathcal{O}$. The alignment graph provides an integrated view on top of the historical semantic descriptions $\delta_T$. Similar to a semantic model, both class and data nodes are contained in $\mathcal{G}_\mathcal{O}$. Note that the alignment graph is weighted by a weighting function $w_\mathcal{O} : \mathcal{E}_\mathcal{O} \mapsto R$ so that edges inferred from the ontology have higher weights than the edges shown in the known semantic models. The details of the algorithm and weighting function are illustrated in \cite{ref_article4}. Then, the top-$\sigma$ candidate semantic models are acquired from the learned semantic types and alignment graph by solving the Steiner Tree Problem(STP) \cite{ref_article4,ref_article11}. Given a graph $G=(V,E)$ and a subset of its nodes $T \subseteq V$, a Steiner Tree $G_s=(V_s,E_s)$ ($T \subseteq V_s \subseteq V$ and $E_s \subseteq E$) is a subtree of $G$ which contains all the nodes in $T$ and may include extra nodes from $V$ to guarantee the connectedness. The Steiner Tree Problem (STP) can be described as finding the Steiner Tree which has the minimum sum of the weights of the edges in $E_s$ with given graph $G$ and a weight function $w_f : E \mapsto \mathcal{R}$ \cite{ref_article12}. For leveraging a STP to formulate the \verb|Rel2Onto| schema mapping problem for a new source $s^*$, we apply the method proposed in \cite{ref_article4} and build the alignment graph $\mathcal{I}_\mathcal{O}^{\mathcal{S}^*}=(\mathcal{V}_\mathcal{O}^{\mathcal{S}^*},\mathcal{E}_\mathcal{O}^{\mathcal{S}^*})$, where $\mathcal{V}_\mathcal{O}^{\mathcal{S}^*}=(\mathcal{V}_\mathcal{O},\mathcal{A}_\mathcal{S^*})$. The set of edges $\mathcal{E}_\mathcal{O}^{\mathcal{S}^*}$ consists of $\mathcal{E}_\mathcal{O}$ and $\mathcal{M}_\mathcal{O}^{\mathcal{S}^*}$ (i.e., $\mathcal{E}_\mathcal{O}^{\mathcal{S}^*}=\mathcal{E}_\mathcal{O} \cup \mathcal{M}_\mathcal{O}^{\mathcal{S}^*}$), where $\mathcal{E}_\mathcal{O}$ is the set of all edges in the alignment graph, and $\mathcal{M}_\mathcal{O}^{\mathcal{S}^*}$ represents the edges which connect each attribute of $s^*$ to the nodes in the alignment graph induced by the semantic types (i.e., the set of nodes in $\mathcal{D}_{\mathcal{G}_\mathcal{O}}$). For example, in Figure~\ref{seed}, the blue-colored dashed lines are the set $\mathcal{M}_\mathcal{O}^{\mathcal{S}^*}$. Next, a weighting function $w_\mathcal{I} : E \mapsto \mathcal{R}^+$ is associated with the alignment graph. Here, we weight the edges of the alignment graph by the weighting function in \cite{ref_article4}. After weighting edges, we use an approximation algorithm of STP, such as the BANKS algorithm \cite{ref_article14}, to build a set of subgraphs $T^*=(V^*,E^*)$ of the alignment graph $\mathcal{I}_\mathcal{O}^{\mathcal{S}^*}$ for the new source $s^*$ . The output of the Steiner tree algorithm is the top-$\sigma$ candidate semantic models. Since our approach attempts to amend one partially correct semantic model, we only select the first candidate semantic model cm with the lowest weight as the seed model for further amending.

{Figure~\ref{seed} shows the seed model of the data source CB generated by the method in \cite{ref_article4} using the known semantic model of source NPG and source GT. This seed model will be used as the start of amending in the next sub-section.}

\subsection{Amending the Seed Model}
Compared with the gold semantic model, some substructures may be missing, and some wrong relationships may appear in the seed model. For example, as Figure~\ref{seed} shows, in the seed model of CB, the red-colored relationships,  {i.e.,} (\verb|E22_Man-Made_Object1|, \verb|P130i_features_are_also_found_on|, \verb|E57_Material1|), {etc.,} are improper and should not appear in the semantic model. To improve the quality of the seed model, these incorrect relationships must be eliminated from seed models. In the meantime, compared to the gold standard model, the seed model lacks three entities (which are in the white background), {i.e.,} \verb|E52_Time-Span1|, \verb|E52_Time-Span2|, \verb|E55_Type1| and their corresponding incoming links.

In this section, we present a way to modify the seed model to move or remove faulty relationships and add missing relationships to the model. Here, the meaning of moving an erroneous relationship is to attach the relationship to a node in the graph. We propose two approaches to remove or move potentially incorrect relationships. First, we use machine learning techniques to distinguish some ambiguous relationships based on the data source. Then, some incorrect substructures of cm can be detected through matching model fragments in a knowledge graph. After removing or moving incorrect relationships, we add potentially missing substructures of cm by using a modified frequent subgraph mining algorithm. As a result, a high-quality semantic model is obtained.

\subsubsection{Move Incorrect Relationships}

One entity may be linked to multiple other entities by various relationships. For example, the entity \verb|E52_Time-Span| can be linked with \verb|E12_Production|, \verb|E8_Acquisition|, \verb|E67_Birth,|, and \verb|E69_Death| by relationship \verb|P4_has_time-span| in the gold semantic models of the \verb|museum-crm| dataset. We define such an entity as an ambiguous entity and the attributes it labeled as ambiguous attributes. In the candidate model generated by the Steiner Tree algorithm, some elements may be wrong because of multiple possible relationships. For example, as Figure \ref{seed} shows, the relationships \verb|P4_has_time-span| from the entity \verb|E12_Production| to \verb|E52_Time-Span2| and \verb|E52_Time-Span3| are incorrect. In fact, these entities should be linked with \verb|E67_Birth| and \verb|E69_Death| by relationship \verb|P4_has_time-span| respectively. In this section, we put forward a machine learning method to move such ambiguous relationships. We treat the problem of distinguishing ambiguous relationships as a multi-category classification problem. Similarity metrics are used as features of the learning matched function to determine whether different ambiguous attributes have the same relationship and thereby infer the correct links.

Our method of removing incorrect relationships is summarized in the following: (1) For training data sources, we gather all ambiguous attributes (their corresponding ambiguous relationships are known), extract several features and train a decision tree model \cite{ref_article28}; (2) For a new data source, we find all the ambiguous attributes and use the trained decision tree model to determine the correct linking position; and (3) We move the relationships according to the predicted result.

We used the following candidate features, including attribute name similarity, value similarity, distribution similarity, and histogram similarity to a decision tree model. Besides these, we also use an external \emph{knowledge base} to generate additional features. We briefly describe these similarities in the following: (1) {Attribute Name Similarity:} 
 Usually, there is a \emph{title} for each column of a structural tabular data source such as a Web table or spreadsheet. We treat these headings as attribute names and use them to compare the similarities between attribute names and entity names. The similarity may infer the correct relationship to which the entity is linked. For example, if an attribute is named \textit{birthDate}, its labeled entity should be \verb|E52_Time-Span| link with entity \verb|E67_Birth| rather than entity \verb|E69_Death|. (2) Value Similarity: Value similarity is the most commonly used similarity measure, which has been used in various matching systems. Since the same semantic types usually contain similar values, value similarity plays a significant role in recognizing attributes labeled by the identical semantic types. In our method, two different value similarity metrics are applied, {i.e.}, Jaccard similarity \cite{ref_article25} and TF-IDF cosine similarity \cite{ref_article25} for computing the value similarity of textual data. (3) Distribution Similarity: For numeric data, value similarity is always ineffective to distinguish semantic types because they always have the similar value range. However, their distribution of values may be different because they have different potential meanings. Therefore, we use statistical hypothesis testing as one of the similarity measures to analyze the distribution of values in attributes. We also used the Kolmogorov--Smirnov test (KS test) \cite{ref_article26} as one of the similarity metrics. (4) Histogram Similarity: Histogram similarity calculates value histograms in textual attributes and compares their histograms. The statistical hypothesis test for the histograms is the Mann--Whitney test (MW test) \cite{ref_article26}. In our method, the MW test is used for computing the histogram similarity, considering it calculates the distribution distance based on medians. (5) External Knowledge Base: To improve the accuracy of our approach further, we used an external knowledge base as a candidate feature. In the cultural heritage community, the Getty Union List of Artist Names (ULAN) {\url{https://www.getty.edu/research/tools/vocabularies/ulan }}(14 December 2022) {is an} 
 authoritative reference dataset containing over 650,000 names of over 160,000 artists. For the museum datasets, some ambiguous attributes can be distinguished by retrieving the information from ULAN. For example, we can validate the information of biography of artists by comparing the information with an attribute labeled by \verb|E52_Time-Span| with ULAN to determine if the information with \verb|E52_Time-Span| really represents the birth date of artists.

Our approach to distinguishing ambiguous attributes is stated in detail as follows. Given an ambiguous entity, where the number of possible links is $k>1$,
we randomly select $k$ attributes \{$a_{r1}$, $a_{r2}$, ..., $a_{rn}$\} from the training attributes as reference attributes. The reference attributes should contain all possible relationships. For other training attributes \{$a_1$, $a_2$, $a_3$, ... $a_n$\}, we compute multidimensional feature vectors $f_i$ by comparing against all reference attributes and searching the external knowledge base. During training, we label each $f_i$ as a number $j$ (ranging from 1 to $k$), where $j$ means the $jth$ possible relationships. Given a new ambiguous attribute $a_0$, we compute the feature $f_0$ and use the learned tree model to label $f_0$ as a number ranging from 1 to $k$. Unlike \cite{ref_article7}, for each attribute, we compute only one feature vector by comparing it with all the reference attributes. If the label of $f_0$ is $m$, it means that $a_0$ is linked by the $mth$ possible relationships. If the predicted result is not consistent with the relationship in the seed model, we move the relationship according to the predicted result. 

\begin{figure}[H]
\begin{adjustwidth}{-\extralength}{0cm}
\centering
\includegraphics[width=1.1\textwidth]{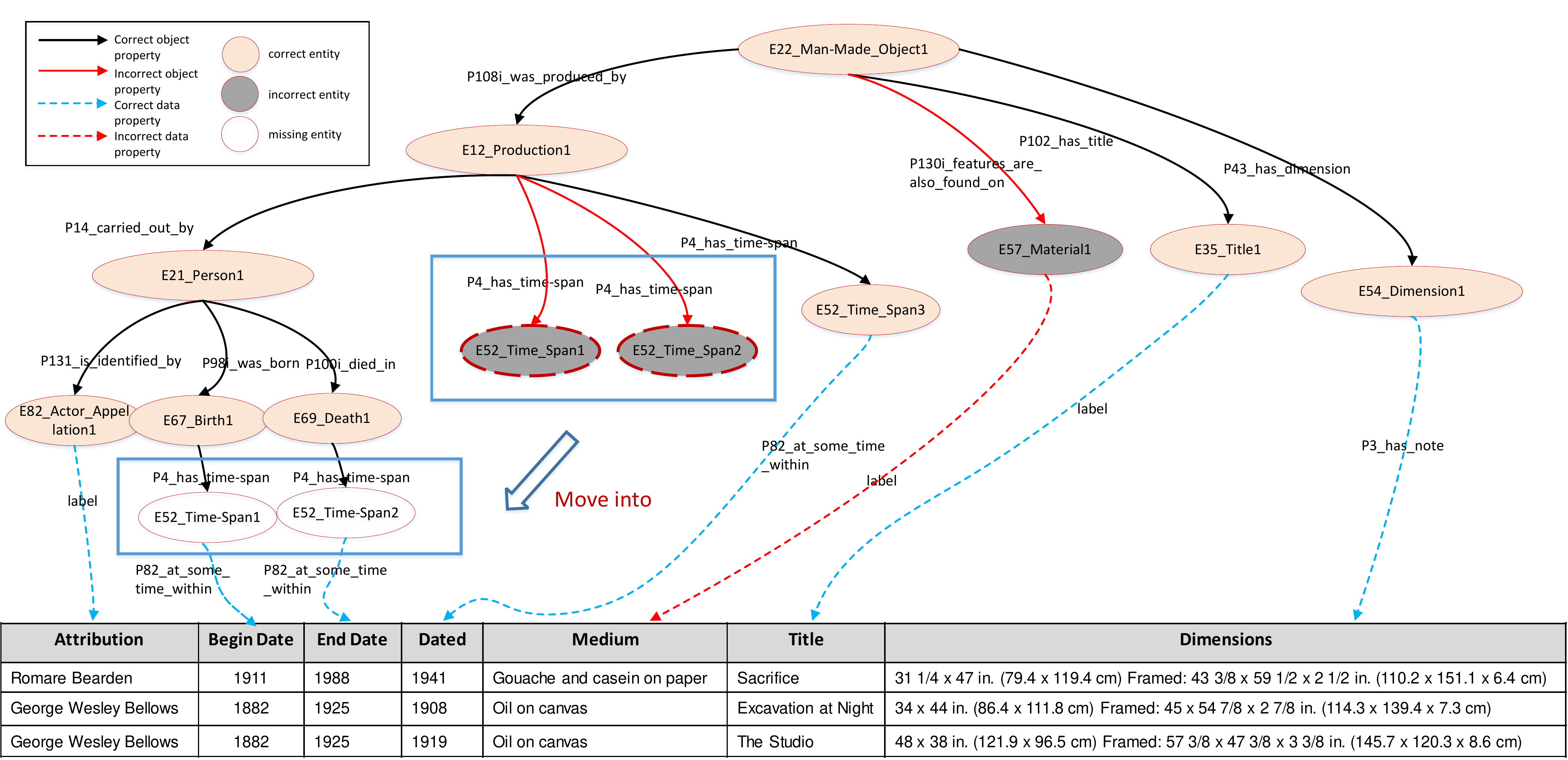}
\end{adjustwidth}
\caption{The change of seed model of CB after moving ambiguous relationships.} 
\label{seed model 1}
\end{figure}

For the seed model of CB, as Figure \ref{seed} shows, the corresponding relationships of ambiguous attributes \verb|Begin Date|, and \verb|Death Date| are (\verb|E12_Production|, \verb|P4_has_time-span|, \verb|E52_Time-Span|). These are wrong predictions by the Steiner Tree algorithm. {Figure \ref{seed model 1} shows the changes after moving the ambiguous relationships.} Through our method, the correct relationships of these attributes can be predicted, and the incorrect relationships are moved into the correct linking position, i.e., (\verb|E67_Birth|, \verb|P4_has_time-span| , \verb|E52_Time-Span1|) for \verb|Begin Date| and (\verb|E69_Death|, \verb|P4_has_time-span| , \verb|E52_Time-Span2|) for \verb|End Date|.

\subsubsection{Remove Incorrect Relationships}

 There may still be incorrect substructures in the candidate model even though we move incorrect relationships. We can leverage the knowledge graph as \emph{prior} knowledge to identify and remove them. The underlying idea of our method is that if incorrect relationships are included in $sd$, then $sd$ has no isomorphic matches in the knowledge graph. If we remove incorrect substructures, the amended model must be a subgraph of the knowledge graph. Algorithm~\ref{alg:removeIncorrectRel} shows how we remove incorrect relationships using a knowledge graph. The input is a seed model $sd$ and a knowledge graph $\mathcal{G}$. First, we remove all the edges in cm that do not appear in the knowledge graph (Lines 4-8).  (\verb|E22_Man-Made_Object|, \verb|P130i_features_are_also_found_on|, \verb|E57_Material|) . Obviously, with this relationship, $sd$ cannot be matched to any occurrence in $\mathcal{G}$.  This type of incorrect substructure is easily detected and removed. However, some incorrect substructures may appear in the knowledge graph but not occur in the gold model of cm. For these parts, we calculate the maximum common subgraph (MCS) \cite{ref_article13} between $\mathcal{G}$ and $sd$ to find a subgraph of $sd$ which is subgraph-isomorphic to $\mathcal{G}$ with maximum nodes. The substructures which do not appear in the $mcs$ are removed (Lines 9-11). At this point, all the incorrect substructures of $sd$ are removed. If we remove an incorrect relationship, the corresponding attribute is temporarily unlabeled by any entities. We denote such an attribute as an isolated column $isoCols$. After removing incorrect relationships, the algorithm iterates over all the columns of $sd$ and returns all the isolated columns that are not annotated by entities. 
 
 {Figure~\ref{seed model 2} shows the change after running Algorithm~\ref{alg:removeIncorrectRel}.  For the seed model of CB, the relationship}  (\verb|E22_Man-Made_Object|, \verb|P130i_features_are_also_found_on|, \verb|E57_Material|) {does not exist in the knowledge graph. Therefore, this incorrect relationship is removed during the process in Lines 4-8 of  Algorithm~\ref{alg:removeIncorrectRel}.}  \verb|Medium| {is the isolated column after removing this incorrect relationships because it is not annotated by any semantic types.} Since the annotation of the isolate column \verb|Medium| is missing, the seed model is an incomplete model and needs to be complemented with the missing substructures. \color{black} In the next step, we use a modified frequent subgraph mining algorithm {for the completion of the seed model.}

\subsubsection{Add Missing Substructures}
In the seed model, some relationships may be missing. To enhance the integrity of the seed model, we need to add extra substructures. Note that there must exist data types in the data source that the added substructures can match. For instance, as shown in Figure~\ref{cb_data source and its semantic model}, if some attributes in the data are date or time matching \verb|E67_Birth1| , it may be reasonable to amend the model and choose entity \verb|E52_Time-Span1| to match these attributes.

Each column of the new data source may correspond to several candidate semantic types, and every semantic type has an ascribed confidence score computed during the initial semantic labeling step. However, the semantic type with the highest confidence score may not capture the correct semantics of the source attribute. For example, the correct semantic type of column \verb|Medium| of data source CB is \verb|E55_Type|, while its first learned semantic type is \verb|E57_Material|. Accordingly, we consider multiple candidate semantic types for each column of the source when semantic models are constructed.

Considering multiple semantic types may make our method less efficient in constructing a semantic model of a source. Hence, we apply several \emph{heuristics} to increase the effectiveness of our method in our algorithm. Algorithm \ref{alg:reduceSemanticTypes} is used to check and reduce candidate semantic types of an isolated column based on their confidence score and a knowledge graph. First, we denote $\eta$ as the ratio between the confidence score of the first ranked candidate type and the second candidate type (Line 4). However, $\eta$ might not be preset  high enough to avoid the consideration of too many candidate semantic types. {We empirically set three  as the threshold value of $\eta$ based on the experimental results in Table \ref{eta}.} If $\eta$ is larger than three, we only select the first candidate type as the final candidate type (Lines 5-8). Second, we remove all the candidate types whose confidence scores are lower than 0.05 of a column (Lines 10-13).  Finally, some candidate semantic types with high confidence scores can be quickly excluded by searching within the knowledge graph using the subgraph matching technique. For the candidate semantic type $ct$ with a relatively high confidence score (larger than 0.05), we attempted to search all the paths $paths$ in $\mathcal{G}$ that connect $ct$ into candidate model $sd$ after removing a presumed incorrect edge. If no such graph connecting a path in $paths$ into $sd$ is subgraph-isomorphic to the knowledge graph, we remove this candidate type (Lines 14-25). A VF2 algorithm \cite{ref_article24} is used for checking subgraph isomorphism. This process aims to find the entities in $sd$ that do not co-occur with other substructures in the knowledge graph. If such entities are considered as candidate semantic types, the frequent subgraph mining algorithm may return none of the resulting models while costing much time.  
After running Algorithm \ref{alg:reduceSemanticTypes}, some candidate types are removed, which prunes large portions of the search space. 

\begin{table}[H]
\caption{{Counts} 
 of mislabeled attributes in three datasets in a different range of $\eta$.}

\begin{tabularx}{\textwidth}{CCCCCCC}

\toprule
\textbf{Datasets} & \boldmath {$\eta > 1$} & \boldmath {$\eta > 2$} & \boldmath {$\eta > 3$} & \boldmath {$\eta > 4$} & \boldmath {$\eta > 5$} & \boldmath {$\eta > 6$} \\ \midrule
$ds_{edm}$ & 30 & 10 & 4 & 3 & 2 & 1 \\
$ds_{crm}$ & 34 & 7 & 2 & 1 & 0 & 0 \\
$ds_{weapon}$ & 20 & 9 & 7 & 7 & 6 & 6 \\ \bottomrule
\label{eta}
\end{tabularx}
\end{table}

\vspace{-12pt}

{In the seed model of CB,  for the isolated column }  \verb|Medium|, {the confidence scores of the last two candidate semantic types are lower than 0.05, so we filter them out in candidate types.} For the incorrect type \verb|E57_Material| of attribute \verb|Medium|,  there is only one possible relationship connecting it into $sd$, {i.e.,} (\verb|E22_Man-Made_Object|, \verb|P45_consists_of|, \verb|E57_Material|). However, if we try to add this relationship in $sd$, we find the intermediate model is not subgraph-isomorphic to $\mathcal{G}$. So the candidate semantic type \verb|E57_Material| might be incorrect, and we can eliminate it from consideration. After reducing the candidate semantic types, there is only one candidate semantic type {i.e.,} \verb|E55_Type1| for the column \verb|Medium|. \color{black}

\clearpage

\begin{algorithm}[!h]

\BlankLine
\footnotesize\textbf{Algorithm} removeIncorrectRel(Seed Model $sd$, Knowledge Graph $\mathcal{G}$)

\footnotesize\Begin{\footnotesize
    $isoCols \Leftarrow \phi$\;
	\ForEach{$e \in sd.edges()$}{
		\If{$e$ not appear in $\mathcal{G}$}{
            remove $e$ in $sd$\;
		}
	}
	\If{$sd~is~not~subgraph~isomorphic~to~\mathcal{G}$}{
	    $sd \Leftarrow MCS(\mathcal{G}, sd)$
	}
    \ForEach{$col \in sd.columns()$}{
        \If{$col~not~annotated~by~an~entity$}{
            $isoCols \Leftarrow result \cup col$
        }
    }

	\textbf{return} $isoCols$\;
}
\caption{Algorithm for removing incorrect relationships}\label{alg:removeIncorrectRel}
\footnotesize\end{algorithm}

\begin{algorithm}[!h]

\BlankLine
\footnotesize\textbf{Algorithm} reduceSemanticTypes(Seed Model $sd$, Knowledge Graph $\mathcal{G}$, isolated column $isoCol$)

\footnotesize\Begin{\footnotesize
    Let $cTypes$ be the top-4 candidate types of $isoCol$\;
    $\eta \Leftarrow cTypes[0].score / cTypes[1].score$\;
    \If{$\eta > 3$}{
        select the first candidate type\;
    }
    
	\ForEach{$ct \in cTypes$}{
	    \If{$ct.score < 0.05$}{
	       $cTypes.remove(ct)$\;
	       \textbf{continue}\;
	    }
        Let $paths$ be all the possible edge paths that connect $ct$ into $sd$\;
        $isIncorrectType$ = \textbf{true}\;
        \ForEach{$path \in paths$}{
            $tmpModel \Leftarrow sd.addPath(path)$\;
            \If{$isSubgraphIsomorphic(\mathcal{G}, tmpModel)$}{
                $isIncorrectType$ = \textbf{false}\;
                \textbf{break}\;
            }
        }
        \If{$isIncorrectType$}{
            $cTypes.remove(ct)$\;
        }
        
	}
	
}
\caption{Algorithm of reducing candidate semantic types for an isolated column}\label{alg:reduceSemanticTypes}
\footnotesize\end{algorithm}

Algorithm \verb|addMissingSubStructures| (Algorithm~\ref{alg:addMissingSubStructure}) is proposed to search and add deleted substructures of imperfect seed semantic model \verb|sd|. First, we use Algorithm~\ref{alg:removeIncorrectRel} to remove incorrect substructures and obtain $isoCols$ of $sd$ (Line 4). If the attribute set $isoCols$ is empty, the algorithm returns $sd$ as the result (Lines 5-7). In this case, none of the incorrect relationships is detected by Algorithm \ref{alg:removeIncorrectRel}. The seed model might be the correct model. Then, we iterate over the set $isoCols$, and for each attribute $isoCol$, we use Algorithm \ref{alg:reduceSemanticTypes} to reduce its candidate semantic types (Lines 8-10). After these steps, we enumerate all the possible combinations of candidate semantic types of $isoCols$ and run the modified frequent subgraph mining algorithm (Lines 13-21). Unlike the traditional algorithm, we abandon the parameter frequency threshold $\tau$ for the uncertainty of the frequency threshold of the correct model. Without the limitation on frequency, the efficiency of frequent subgraph mining algorithms may decrease. Accordingly, we propose a set of pruning strategies for speeding up our algorithm. These pruning strategies will be introduced below. The output of Algorithm \ref{alg:addMissingSubStructure} is the set $result$, which stores all frequent subgraphs as complementary semantic models.

\begin{figure}[H]
\begin{adjustwidth}{-\extralength}{0cm}
\centering
\includegraphics[width=1.1\textwidth]{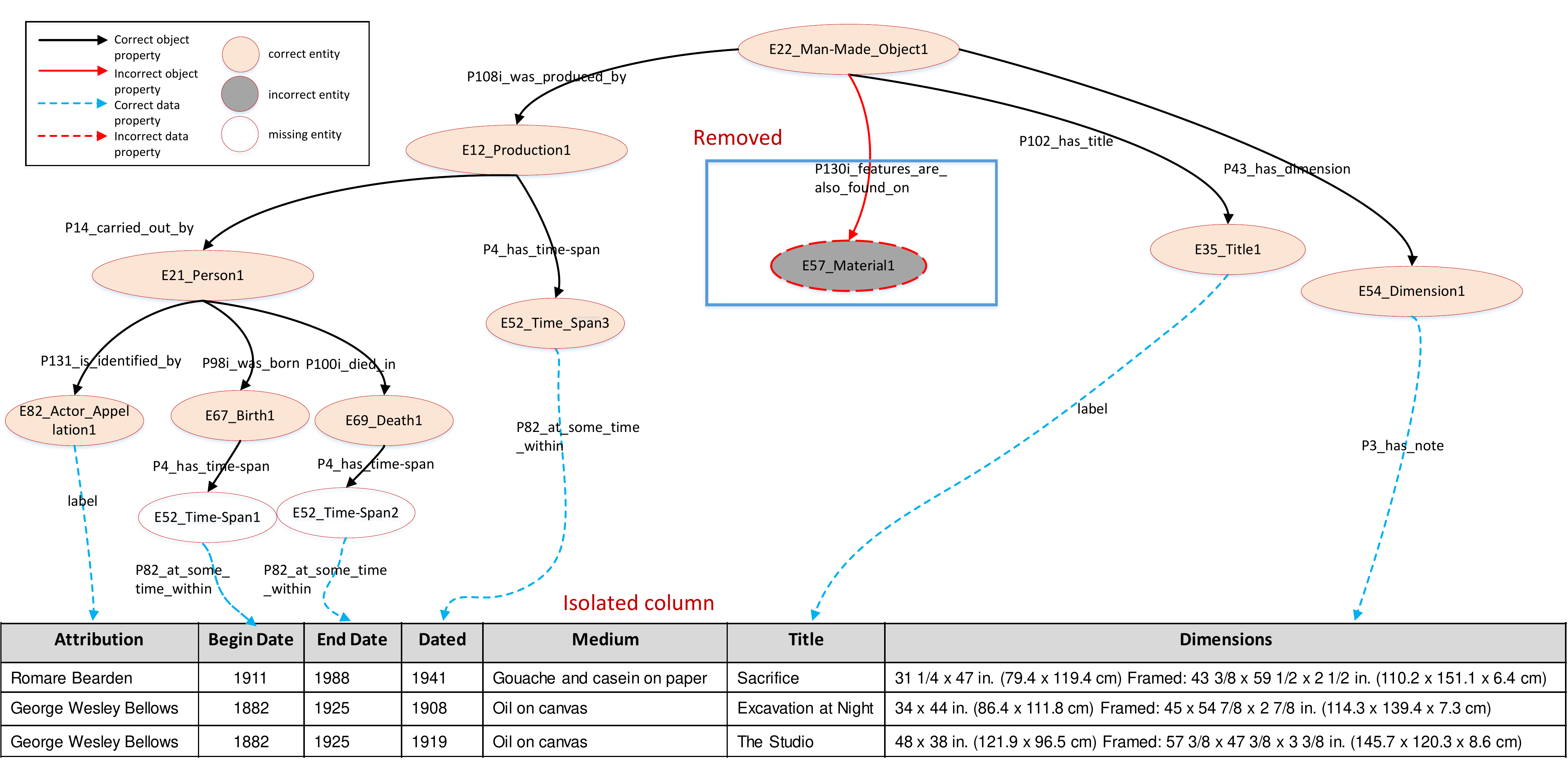}
\end{adjustwidth}
\caption{The change of seed model of CB after removing incorrect relationships.} 
\label{seed model 2}
\end{figure}
\vspace{-6pt}

\begin{algorithm}[!h]

\BlankLine
\footnotesize\textbf{Algorithm} addMissingSubStructures(Seed Model $sd$, Knowledge Graph $\mathcal{G}$, Constraint Map cm)

\footnotesize\Begin{\footnotesize
    $result \Leftarrow \phi$\;
	$isoCols \Leftarrow removeIncorrectRel(sd, \mathcal{G})$\;
	\If{$isoCols == \phi$}{
	    \textbf{return $sd$}\;
	}
    \ForEach{$isoCol \in isoCols$}{
        $reduceSemanticTypes(sd, \mathcal{G}, isoCol)$\;
    }
    Let $types$ be the set of the combinations of candidate types of $isoCols$\;
	Let $Edges$ be the set of edges of in $\mathcal{G}$\;
    \ForEach {$newNodes \in types$} {
		\ForEach{$e \in Edges$}{
				\If{$e$ is the edge of sd}{
						$result \Leftarrow result \cup subgraphExtension(\mathcal{G},e,newNodes,sd,cm)$\;
						Remove $e$ from $Edges$\;
				}
		}
    }
    \textbf{return} $result$\;
}

\caption{Algorithm for repairing the seed model}\label{alg:addMissingSubStructure}
\footnotesize\end{algorithm}

\begin{algorithm}[!h]
\BlankLine
\footnotesize\textbf{function} \textbf{subgraphExtension}(Knowledge Graph $\mathcal{G}$, Incomplete semantic model $\mathcal{S}$, Semantic Labels $newNodes$, Seed Model $sd$, Constraint Map cm)

\footnotesize\Begin{\footnotesize
    $freqs \Leftarrow [0]$\;
		$cm=HashMap<node,value>$ 
		$//cm~keys:~entity~types~that~appear~in~the~semantic\newline~model~to~be~repaired$; 	$cm~values:~maximum~occurrence~of~the~key~in~the\newline~semantic~model~to~be~repaired$\;
		 \ForEach{$e \in Edges$ and $node~n \in \mathcal{S}$}{
				\If{$e$ can be used to extend $n$ and $e$ is a valid edge}{
					  Let $ext$ be the extension of $\mathcal{S}$ with $e$\;
						\If{$ext$ covers all nodes in $sd$ and $newNodes$ and $sd$ is a subgraph of $ext$}{
								$result \Leftarrow result \cup ext$\;
								\textbf{return} $result$\;
						}
						check every key $et$ in Constraint Map cm\;
						\If{$count(et,ext) > cm[et]$}{
								\textbf{continue}\;
						}
						\ForEach{$e \in sd.edges$}{
						    \If{$e.source \in ext.nodes~and~e.target \in ext.nodes~and~e \notin ext.edges$}{
						        \textbf{continue}\;
						    }
						}\color{black}
						\If{$freq(ext,G) > min(freqs)$}{
								$freqs \Leftarrow freqs \cup freq(ext,G)$\;
								$sort(freqs)$\;
								remove the $min$ element in $freqs$\;
								\If{$ext$ has not been generated before}{
										$result \Leftarrow result \cup$ \textbf{subgraphExtension}($\mathcal{G}$,$ext$,$newNodes$,$sd$,cm)\;
								}
						}
				}
	  }
       \textbf{return} $result$
}
\caption{Algorithm of subgraph extension}\label{alg:subgraphExtension}
\footnotesize\end{algorithm}

As the sub-function of Algorithm~\ref{alg:addMissingSubStructure}, Algorithm \verb|subgraphExtension| (Algorithm~\ref{alg:subgraphExtension}) is used to search and add missing substructures from a specific imperfect semantic model $\mathcal{S}$. The inputs of Algorithm~\ref{alg:subgraphExtension} are a seed model \verb|sd|, a knowledge graph $\mathcal{G}$, an incomplete semantic model $\mathcal{S}$, a constraint map cm, and the candidate semantic types $newNodes$ that the seed model may link with. The \textit{grow-and-store} strategy is adopted in the algorithm \verb|subgraphExtension| for recursively mining the top-$\sigma$ frequent subgraphs of $\mathcal{G}$. In the meantime, Algorithm~\ref{alg:addMissingSubStructure} can ensure the coverage between the seed model \verb|sd| and all of the candidate semantic types $newNodes$ of source attributes in the mined top-$\sigma$ frequent substructures. Here, our \textit{heuristic} is that if the frequency of an amended semantic model is higher than a certain threshold in the knowledge graph, it indicates that the  model could be correct. Further, the higher the frequency of an amended model in the knowledge graph, the more likely it is to be a reasonable semantic model. The output of the algorithm \verb|subGraphExtension| is the top-$\sigma$ semantic models with complemented missing substructures.

The inputs of Algorithm~\ref{alg:subgraphExtension} are illustrated as below. \verb|Semantic Labels| is a candidate combination of candidate semantic types of the source attributes that does not occur in the seed model of a new data source.  \verb|Constraint Map| is a HashMap which constraints the maximum count of entities for each semantic type that may appear in a correct model. Generally, \verb|Constraint Map| is prescribed through domain expertise. For instance, as Figure~\ref{cb_data source and its semantic model} shows, in the semantic model of the CB, the \verb|Constraint Map| $\{<\verb|E52_Time-Span|,3>,<\verb|E35_Title|,1>\}$ limits the count of entities for the semantic types \verb|E52_Time-Span|, and \verb|E35_Title| must be less than three and one, respectively. The intermediate variable $freqs$ is a sorted integer list used to record the top-$\sigma$ frequencies of all subgraphs throughout the algorithm process.

For each relationship $e$ in knowledge graph $\mathcal{G}$ and a semantic type $n$ in the incomplete model $\mathcal{S}$, first Algorithm~\ref{alg:subgraphExtension} validated if $e$ can be linked with the semantic type $n$ and whether $e$ is a valid edge in the knowledge graph (Line 6). For instance, assume that entities \verb|E22_Man-Made_Object1| and \verb|E35_Title1| and a relationship \verb|P102_has_title| occur in an initial subgraph. Suppose $n$ is semantic type \verb|E22_Man-Made_Object1| in $S$ and $e$ is object property \verb|P43_has_dimension|. Since $\mathcal{G}$ contains the triple (\verb|E22_Man-Made_Object1|, \verb|P43_has_dimension|, \verb|E54_Dimension1|), it implies that \verb|P43_has_dimension| is effective and can be used to link with \verb|E22_Man-Made-Object1|. Let $ext$ be the extension of $\mathcal{S}$ with $e$ (Line 7). During the run of Algorithm~\ref{alg:subgraphExtension}, when all the nodes and links in the seed model $sd$ and candidate semantic types $newNodes$ are covered in $ext$, the $ext$ is the most possibly plausible semantic model which correctly captures the semantics of the data source. Next, $ext$ is merged into the set $result$, and the algorithm ceases and returns $result$ (Lines 8-11). The set $result$ is used to store the top-$\sigma$ semantic models as the output of the algorithm \verb|subGraphExtension|. As  Algorithm \ref{alg:addMissingSubStructure} demonstrates, the set $newNodes$ represents one possible combination of all candidate semantic types of $isoCols$, $newNodes$ may be erroneous to annotate semantic types for $isoCols$. Algorithm \ref{alg:subgraphExtension} may return none, which indicates that there are some incorrect candidate types in the $newNodes$.
 
 We apply some pruning strategies to improve the efficiency of the algorithm. In order to prevent the recurrence of a given entity type $et$ in the model, one of the pruning strategies we employ is to leverage \verb|Constraint Map| to limit the searching space. We check each entity $et$ in \verb|Constraint Map| cm. If there is an entity type $et$ that appears in $ext$ exceeding $cm(et)$ times, the further search for current $ext$ is ceased (Lines 13-15). The Function \verb|count()| is used to calculate the number of appearances of $et$ in the intermediate subgraph $ext$. 

We reduce the search space by leveraging the structure of $sd$. We iterate over the edge of $sd$, and if there is an edge $e$ whose source node and target node exist in $ext$ but $e$ does not exist in $ext$, we stop searching $ext$ (Lines 16-20). For example, in the seed model of CB, entity \verb|E22_Man-Made_Object| and entity \verb|E12_Production| are connected by relationship \verb|P108i_was_produced_by|. For an intermediate substructure $ext$ containing these entities, if there is no link between the two entities, $ext$ can not be extended to the model which contains the relationship \verb|P108i_was_produced_by|. It is vain to further search substructure $ext$. In practice, a large amount of search space can be reduced by using this pruning strategy. For the seed model of CB, the running time is about 15 s, while without this optimization, the respective running time is about 5 min.

Next, we use a Minimum Image-based Metric based on a subgraph matching algorithm \cite{ref_article15} (Line 21) in the function $freq(ext,G)$ to calculate the frequencies of $ext$ occurring in $G$. All the substructure $ext$ in which the frequency is lower than the minimum frequency of $freqs$ will be discarded. The intuition behind this pruning strategy is that the correct semantic model may be a subgraph with higher frequency in the knowledge graph. During the process of search, only the top-$\sigma$ frequent subgraphs are retained as candidate semantic models in the frequency-based pruning strategy.

In Algorithm~\ref{alg:subgraphExtension}, during the whole search, top-$\sigma$ frequencies of all subgraphs are stored in an integer list $freqs$ (line 22) in which the length is $\sigma$. Here, $freqs$ merges the frequency of $ext$ ($freq(ext,G)$) (Line 22) if the frequency of $ext$ is higher than the minimum value of the current $freqs$. Then, Algorithm~\ref{alg:subgraphExtension} sorts $freqs$ (Line 23) and removes the minimum value in the sorted $freqs$(Line 24). 

Thereafter, for removing duplicate models, Algorithm~\ref{alg:subgraphExtension} checks if $ext$ has been searched in the previous procedure. We exploit the \textit{canonical code} method proposed in \cite{ref_article6} to detect the duplicate models (Line 25). The algorithm \verb|subgraphExtension| is recursively executed (Line 26) for further search if the substructure $ext$ has not been generated before.

 The incomplete model of CB before adding missing substructures is shown in Figure~\ref{seed model 2} (wrong entity \verb|E57_Material1| and its relationship were removed). After reducing the candidate semantic types, \verb|E55_Type1| is the only candidate type for isolated column \verb|Medium|. Through the process of our modified subgraph mining algorithm, the missing relations (\verb|E12_Productioin1|, \verb|P32_used_general_technique|, \verb|E55_Type1|) can be linked into the incomplete model, and the gold standard model (Figure~\ref{cb_data source and its semantic model}) can be output as a frequent subgraph. Hence, the seed model of CB is amended into a correct semantic model. \color{black}

\section{Evaluation}\label{55}
\subsection{Experimental Setting}

To assess our method, we conducted the experiments on three datasets, i.e., \verb|museum_edm| ($ds_{edm}$), \verb|museum_crm| ($ds_{crm}$), and \verb|wepaon_lod| ($ds_{weapon}$). The datasets $ds_{edm}$ and $ds_{crm}$ both contain 29 different data sources from different art museums in the USA and have different data formats (CSV, XML, and JSON). Nevertheless, two different famous data models are used as museum domain ontology: European
Data Model {(EDM)} 
 {\url{http://pro.europeana.eu/page/edm-documentation}}(accessed on 14 December 2022), and CIDOC Conceptual Reference
Model {(CIDOC-CRM)} 
 {\url{www.cidoc-crm.org}}(accessed on 14 December 2022); $ds_{weapon}$ includes 15 data sources about weapon ads. The ontology of $ds_{weapon}$ is an extension of the schema.org ontology. The background knowledge graphs were constructed by capturing these data sources and mapping them to the corresponding domain ontology. For each specified data source of a dataset, we built a knowledge graph that integrated the data from all of the data sources excluding this one. For example, for the $s_1$ of $ds_{crm}$, the knowledge graph integrates all the data sources, i.e., $s_2$, $s_3$, ..., $s_{29}$, except $s_1$. Table \ref{datasets} lists the details of these three datasets. For facilitating the construction of knowledge graphs, we reconstructed the semantic models of the ground-truth datasets and transformed all the data sources into CSV format. The  datasets, experimental results, and our code are available on {Github}
 {\url{https://github.com/Zaiwen/ModelCorrection}} (accessed on 14 December 2022). Our objective is to assess the effectiveness of our method if only a few known semantic models of similar sources are available. So we use only two or three data sources of the datasets for training and the others for testing. We repeat the process three times and average the results. Our experiments were run on a single machine with an Intel i7 10500 CPU 3.40GHz and 16 GB RAM.

\begin{table}[H]
\caption{The evaluation of datasets.}\label{datasets}
\newcolumntype{C}{>{\centering\arraybackslash}X}
\begin{tabularx}{\textwidth}{lCCC}
\toprule
 & \boldmath {$ds_{edm}$} & \boldmath {$ds_{crm}$} & \boldmath {$ds_{weapon}$} \\ \midrule
\#data sources & 29 & 29 & 15 \\
\#classes in domain ontologies & 120 & 83 & 718 \\
\#properties in domain ontologies & 351 & 270 & 295 \\
\#nodes in the gold-standard models & 409 & 750 & 230 \\
\#data nodes in the gold-standard models & 123 & 362 & 81 \\
\#class nodes in the gold-standard models & 286 & 388 & 149 \\
\#links in the gold-standard models & 380 & 724 & 215 \\ 
\#average entities in knowledge graphs & 55432 & 57558 & 5403 \\
\#average relationships in knowledge graphs & 73722 & 82654 & 6529 \\
\bottomrule
\end{tabularx}
\end{table}

\subsection{Empirical Preliminary Experiments}

In learning the candidate semantic types of an attribute of a data source $s_i$, we use the known semantic models and their corresponding data sources as training data. For example, if we are learning the candidate types of data source $s_i$, the training data are all the data sources \{$s_k|k=1,...,j~and~k\ne i$\}. The semantic labeler we use is {SemanticTyper} \cite{ref_article23}. In this work, we only consider the top four semantic types. Table \ref{mrr} shows the mean reciprocal rank (MRR) \cite{ref_article20} scores of semantic labeling in three datasets.

\begin{table}[H]
\begin{center}
\caption{MRR of semantic labeling.}
\begin{tabularx}{\textwidth}{CCCC}
\toprule
\textbf{Datasets} & \boldmath {$ds_{edm}$} & \boldmath {$ds_{crm}$} & \boldmath {$ds_{weapon}$} \\ \midrule
MRR scores & 0.907 & 0.937 & 0.879 \\ \hline
\end{tabularx}
\label{mrr}
\end{center}
\end{table}

For an attribute, if the confidence score of its first ranked candidate semantic type is much higher than the second candidate type, the first ranked candidate type is assumed to be correct. To reduce the number of possible candidate types, we consider only the first ranked candidate semantic type for such an attribute. Let $\eta$ be the ratio between the confidence score of the first candidate type and the second candidate type. Table~\ref{eta} shows the counts of the mislabeled attributes in different ranges of $\eta$. The counts of mislabeled attributes decrease with increasing $\eta$. However, there are a few mislabeled attributes whose $\eta$ is very large. When $\eta$ is larger than three, the count of mislabeled attributes in three datasets tends to be relatively stable. Therefore, in Algorithm \ref{alg:reduceSemanticTypes}, we only select the first candidate type for the attribute whose $\eta$ is larger than three for the balance between the accuracy of candidate types and the efficiency of our algorithm.

\subsection{Effectiveness of Our Approach}

To evaluate our approach, standard mean reciprocal rank (MRR) \cite{ref_article20} was used. We 
compared the obtained models with the gold semantic model to assess the correctness of them based on precision and recall as in Taheriyan et al. \cite{ref_article4}:

\begin{linenomath}
    \begin{equation}
        precision= \frac{|rel(sm) \cap rel(f^*(sm'))|}{|rel(f^*(sm'))|}
    \end{equation}
\end{linenomath}

\begin{linenomath}
    \begin{equation}
        recall=\frac{|rel(sm) \cap rel(f^*(sm'))|}{|rel(sm)|}
    \end{equation}
\end{linenomath}

\begin{linenomath}
    \begin{equation}
        f^*=\mathop{argmax}\limits_{f}|rel(sm) \cap rel(f(sm'))|
    \end{equation}
\end{linenomath}
where $rel(sm)$ is the set of the triples $(u,e,v)$ of a semantic model $sm$, and $f$ is a mapping function, which maps nodes in $sm'$ to nodes in $sm$.

Table~\ref{result} shows the results of our experiments. Two state-of-the-art semantic modeling systems were compared: \verb|Karma| \cite{ref_article4} and \verb|PGM-SM| \cite{ref_article27}. Among them, our method improves the two baseline methods by an average of 10.68\%, 13.85\%, and 9.08\% on $ds_{edm}$, $ds_{crm}$, and $ds_{weapon}$, respectively. Our approach uses knowledge graphs for amending the incorrect substructures in learned semantic models and realizes noticeable modification, indicating that knowledge graphs are useful prior knowledge to improve the quality of learned semantic models.

\begin{table}[H]
\caption{Performances of our method on $ds_{edm}$, $ds_{crm}$, and $ds_{weapon}$.}
	\begin{adjustwidth}{-\extralength}{0cm}
		\newcolumntype{C}{>{\centering\arraybackslash}X}
		\begin{tabularx}{\fulllength}{ccCcCCcCCcC}
\toprule
\multirow{2.5}{*}{\textbf{Datasets}} & \multirow{2.5}{*}{\textbf{known Models}} & \multicolumn{3}{c}{\textbf{Precision}}   & \multicolumn{3}{c}{\textbf{Recall}}   & \multicolumn{3}{c}{F1} \\ \cmidrule{3-5} \cmidrule{6-8} \cmidrule{9-11} 
 &  & \textbf{Karma} & \textbf{PGM-SM} & \textbf{Ours}   & \textbf{Karma} & \textbf{PGM-SM} & \textbf{Ours}   & \textbf{Karma} & \textbf{PGM-SM} & \textbf{Ours} \\ \midrule
\multirow{2}{*}{$ds_{edm}$} & 2 & 0.864 & 0.791 & {\textbf{0.920}} 
   & 0.846 & 0.754 & \textbf{0.913}   & 0.855 & 0.770 & \textbf{0.917} \\
 & 3 & 0.858 & 0.778 & \textbf{0.920}   & 0.840 & 0.757 & \textbf{0.912}   & 0.849 & 0.765 & \textbf{0.916} \\ \midrule
\multirow{2}{*}{$ds_{crm}$} & 2 & 0.738 & 0.824 & \textbf{0.897}   & 0.703 & 0.721 & \textbf{0.878}   & 0.721 & 0.773 & \textbf{0.888} \\
 & 3 & 0.770 & 0.828 & \textbf{0.908}   & 0.731 & 0.725 & \textbf{0.892}   & 0.751 & 0.777 & \textbf{0.900} \\ \midrule
\multirow{2}{*}{$ds_{weapon}$} & 2 & 0.795 & 0.809 & \textbf{0.870}   & 0.734 & 0.758 & \textbf{0.834}  & 0.765 & 0.784 & \textbf{0.852} \\
 & 3 & 0.837 & 0.805 & \textbf{0.924}   & 0.783 & 0.810 & \textbf{0.902}   & 0.810 & 0.808 & \textbf{0.913} \\ \bottomrule
\end{tabularx}
	\end{adjustwidth}
\label{result}
\end{table}

\begin{figure}[H]
\includegraphics[width=\textwidth]{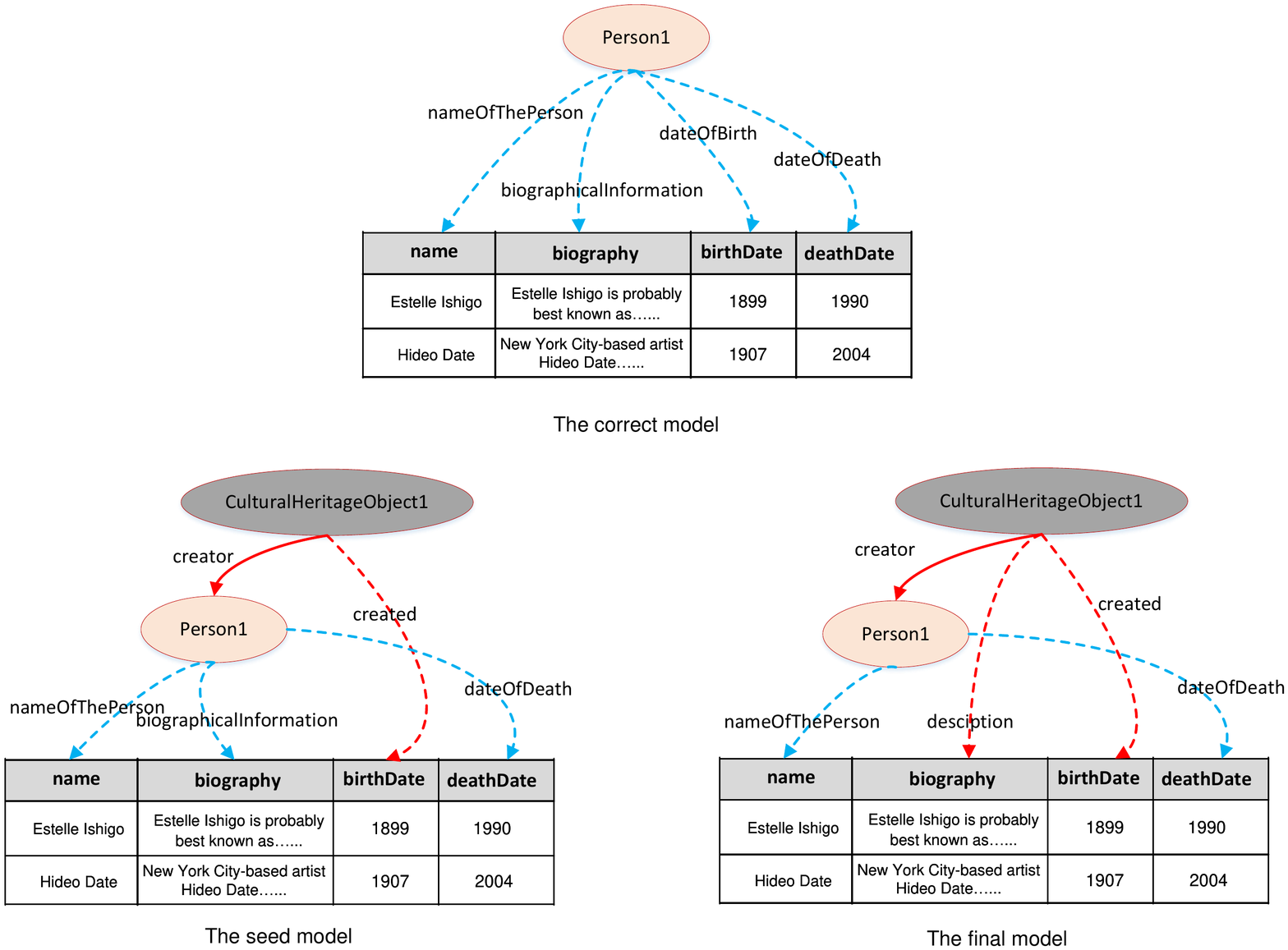}
\caption{The predict result of $s_{9}$ in $ds_{edm}$, using $s_2$ and $s_6$ as training sets.} 
\label{abnormal result}
\end{figure}

While our approach performs well in most cases, the accuracy of the semantic model generated by our method is inferior to the semantic model generated by \verb|KARMA| in some specific cases. For example, the prediction accuracy and recall of the semantic model generated by \verb|KARMA| of $s_9$ are 0.6 and 0.75, respectively, while $s_2$ and $s_6$ are used as the training set of $ds_{edm}$. However, after amending using our approach, the precision and recall of the final model are 0.4 and 0.5, respectively. Figure~\ref{abnormal result} shows the correct model, seed model, and final model of $s_9$ in $ds_{edm}$. Compared to the seed model, our approach moves the correct relationship (\verb|Person|, \verb|biographicalInformation|, \verb|biography|) into the error relationship (\verb|CulturalHeritageObject1|, \verb|description|, \verb|biography|). The incorrect entity \verb|CulturalHeritageObject1| appears in the seed model because of the wrong predicted semantic type of attribute \verb|birthDate|. The relationship (\verb|Person1|, \verb|biographicalInformation|, \verb|biography|) does not co-occur with the entity \verb|CulturalHeritageObject1| in the knowledge graph. During running Algorithm~\ref{alg:removeIncorrectRel}, the relationship (\verb|Person1|, \verb|biographicalInformation|, \verb|biography|) is removed at line 10. This phenomenon indicates that our approach is sensitive to the incorrect semantic labeling result.
Since our approach attempts to improve the seed model generated by the Steiner Tree algorithm, the performance of our approach is highly sensitive to the quality of the seed model. If the seed model differs greatly from the corresponding correct model, our approach may be unable to recover the correct model.

\subsection{Efficiency of Our Approach}

We measured the running time of our method. The results are listed in Table \ref{running time}. Phase I refers to the process of moving and removing the incorrect substructures and phase II is the process of adding missing substructures. The running time of our method is positively correlated with the size of the knowledge graph. Since the size of knowledge graphs in $ds_{weapon}$ is much smaller than that of $ds_{edm}$ and $ds_{crm}$ (Table \ref{datasets}), we can see that the running time is much less. In our approach, the running time mainly depends on graph algorithms whose running time is affected by the size of the knowledge graph and the size of the semantic models. While the size of the knowledge graphs created in different scenarios in our experiment is close, the size of semantic models in $ds_{edm}$ is smaller than that in $ds_{crm}$. Therefore, the running time of phase II of $ds_{edm}$ is less.

\begin{table}[H]

\caption{Average running time of our method on $ds_{edm}$, $ds_{crm}$ and $ds_{weapon}$.}
\begin{tabularx}{\textwidth}{CCC}
\toprule
\textbf{Datasets} & \textbf{Phase I} & \textbf{Phase II} \\ \midrule
$ds_{edm}$ & 45.395s & 13.581s \\
$ds_{crm}$ & 46.782s & 33.543s \\
$ds_{weapon}$ & 4.232s & 2.730s \\ \bottomrule
\end{tabularx}
\label{running time}
\end{table}

\section{Conclusions}\label{66}
In this article, we propose a novel approach for solving the \verb|Rel2Ont| problem by leveraging a knowledge graph as background knowledge. First, we require a partially correct semantic model called seed model by running the Steiner tree algorithm \cite{ref_article14}. We move or remove imperfect relationships in the seed model by using machine learning and a graph matching technique. After eliminating the incorrect substructures, a modified frequent subgraph mining algorithm is applied to search the top-$\sigma$ frequent substructures covering the seed model and the source attributes candidate semantic types from the domain knowledge graph. Our experimental results indicate that we can generate high-quality semantic models even when the known semantic models are lacking and that we can outperform two state-of-the-art semantic modeling systems in terms of the correctness of the resulting models. In the future, we would like to further develop our method in the following areas. Firstly, our decision tree model for distinguishing ambiguous relationships can be enhanced by a feature selection algorithm \cite{ref_article29} to further improve the accuracy. Secondly, we will explore an automatic method for extracting a \verb|Constraint Map| from historical data. {Finally, we will try to extend our approach to those data sources containing a set of relations}.

\vspace{6pt} 



\authorcontributions{ Conceptualization, W.M. and Z.F.; Methodology, J.X.; Software, J.X.; Validation, J.X.; Formal analysis, J.X.; Investigation, J.X.; Resources, H.Z.; Data curation, J.X.; Writing(original draft), J.X.; Writing (review and editing), J.X., W.M. and Z.F.; Visualization, J.X.; Supervision, H.Z., K.H. and Z.F.; Project administration, Z.F.; Funding acquisition, Z.F. All authors have read and agreed to the published version of the manuscript.}

\funding{{ This research project was supported in part by the open funds of the National Key Laboratory of Crop Genetic Improvement under Grant ZK202203, Huzhong Agricultural University, and in part by the Major Project of Hubei Hongshan Laboratory under Grant 2022HSZD031, and in part by the Innovation fund of Chinese Marine Defense Technology Innovation Center under Grant JJ-2021-722-04, and in part by the Fundamental Research Funds for the Chinese Central Universities under Grant 2662020XXQD01, 2662022JC004, and in part by the open funds of State Key Laboratory of Hybrid Rice, Wuhan University. } }




\dataavailability{{ The data that support the findings of this study are openly available in Github at} \url{https://github.com/Zaiwen/ModelCorrection } }

\acknowledgments{This research project was supported in part by the open funds of the National Key Laboratory of Crop Genetic Improvement under Grant ZK202203, Huzhong Agricultural University, and in part by the Major Project of Hubei Hongshan Laboratory under Grant 2022HSZD031, and in part by the Innovation fund of Chinese Marine Defense Technology Innovation Center under Grant JJ-2021-722-04, and in part by the Fundamental Research Funds for the Chinese Central Universities under Grant 2662020XXQD01, 2662022JC004, and in part by the open funds of State Key Laboratory of Hybrid Rice, Wuhan University. Some of the initial work in this article was done when the last author did his post-doctoral research in the ITMS, University of South Australia between 2016 and 2019. We appreciate the valuable suggestion from Markus Stumptner, Georg Grossmann, Wenhao Li, Selasi Kwashie, and Amir Kashefi from ITMS, University of South Australia, and Wangyu Huang from Data-to-Decision CRC. Numerical computations were performed on the Hefei Advanced Computing Center in China.}

\conflictsofinterest{The authors declare no conflict of interest.} 

\begin{adjustwidth}{-\extralength}{0cm}

\reftitle{References}

\end{adjustwidth}
\end{document}